# Quantum Neural Network Restatement of Markov Jump Process

Z. Zarezadeh[1], N. Zarezadeh[2]
[1]E-mail: zakarya.zarezadeh@gmail.com
[2]E-mail: naeim.zarezade@gmail.com

*Abstract—* **Despite the many challenges in exploratory data analysis, artificial neural networks have motivated strong interests in scientists and researchers both in theoretical as well as practical applications. Among sources of such popularity of artificial neural networks the ability of modeling non- linear dynamical systems, generalization, and adaptation possibilities should be mentioned. Despite this, there is still significant debate about the role of various underlying stochastic processes in stabilizing a unique structure for data learning and prediction. One of such obstacles to the theoretical and numerical study of machine intelligent systems is the curse of dimensionality and the sampling from high-dimensional probability distributions. In general, this curse prevents efficient description of states, providing a significant complexity barrier for the system to be efficiently described and studied.**

**Therefore, the complexity of data-driven computational statistics requires the development of new theoretical modeling of the dynamics of such probabilistic processes. In this strand of research, direct treatment and description of such abstract notions of learning theory in terms of quantum information be one of the most favorable candidates. Hence, the subject matter of these articles is devoted to problems of design, adaptation and the formulations of computationally hard problems in terms of quantum mechanical systems. A convenient abstraction of such theory in the Markovian regime is system-bath interaction where the heat bath replaced by an ensemble of random-matrices and the time evolution of the quantum trajectory given by the stochastic master equation. In order to characterize the microscopic description of such dynamics in the language of inferential statistics, covariance matrix estimation of *d*-dimensional Gaussian densities and Bayesian interpretation of eigenvalue problem for dynamical systems is assessed.**
.

***Keywords:*** *Quantum Fisher Information; Lindblad master equation; Density matrix; Bayesian inferences; Quantum information science*

## I. Introduction

IN context of Bayesian data analysis and probabilistic inference the evaluation of high-dimensional integrals is a notorious difficult problem[1-4]. In this respect the state-of-the-art Markov chain Monte Carlo (MCMC) methods used in order to traverse the distances in probability space and asymptotically generate samples from posterior distribution [4-6]. Therefore, the basic idea of MCMC integration is to evaluate an integral by sampling the integrand at points selected at random from a distribution proportional to the integration measure.

In order to encourage the efficient sampling and probability density exploration, Hamiltonian Monte Carlo (HMC) proposal with some limitations have been successfully applied to a large class of inference problems [7]. However, given the structure of the parameter space of statistical models, there is no guiding principle on how HMC sampler should be tuned. Since Riemann manifold Hamiltonian Monte Carlo (RMHMC) scheme automatically adapts its mass matrix via position dependent metric tensor, unfortunately the calculation of derivatives associated with metric tensor and its inverse to be computationally too costly to implement. In this paper, we first review some important properties of stochastic differential equations and then we derive a novel extension of probability space exploration with the corresponding stochastic dynamics. Among these results one notes the theory of random matrices and diffusion equation to evaluate an integral over the unitary group. With a coherent treatment of the subject, we observe that the local statistical behaviour of the energy contours for the probability densities could be simulated by the eigenvalues of a random matrices. Without loss of generality, the remainder of this section assumes that the target density function to sample from follows a Gaussian distribution

$$p(\theta) = (2\pi)^{-n/2}|\Sigma|^{-1/2}\exp[-\frac{1}{2}(x-\mu)^T\Sigma^{-1}(x-\mu)]$$

With mean vector $\mu$, $n \times n$ positive definite covariance matrix $\Sigma$. Under this assumption Fisher information matrix (FIM) reduce to

$$\mathcal{I} = S^T\Sigma^{-1}S$$

With stochastic sensitivity matrix S defined to be

$$S_j(x,t) = \frac{\partial \log p(x,t)}{\partial \theta_j}$$

In order to establish a clear and consistent notation we also define the classical Fisher information as $\mathcal{J}_{\text{CFI}}$ and the quantum analogue of the classical one as $\mathcal{J}_{\text{QFI}}$.

Throughout the numerical experiments we will sketch the performance of proposed methodology which gives a substantial gain over the standard HMC methods.

The outline of the paper is as follows: In Section II we introduce the basic concept of statistical inference and briefly recall the definition of HMC methods, in Section III the general expression of Quantum Fisher information matrix (QFIM) for density matrix $\rho$ obtained hence the mass kernel scheme encoded in density matrix $\rho$ is derived, through the discussion of numerical simulation and implication scheme in Section IV we finally sketch our conclusions.

## II. HAMILTONIAN MONTE CARLO SCHEME

Hamiltonian Monte Carlo (HMC) methods are MCMC methods designed to efficiently sample the posterior density by introducing the auxiliary momentum variable $p$ in one-to-one correspondence with generalized coordinated $q$. Starting with a previous value of $q$ the momentum variable $p$ is generated from a multivariate Gaussian distribution. Let us recall the Hamiltonian version of classical Mechanics in the following setting, consider $N$ particles in $\mathbb{R}^d$ of coordinates $q_i \in \mathbb{R}^d$, masses $m_i$ and momenta $p_i \in \mathbb{R}^d, i = 1, \ldots, N$, interacting in a potential $V: \mathbb{R}^{dN} \to \mathbb{R}, q \mapsto V(q)$. The space $\mathbb{R}^{dN}$ of coordinates $(q_1, q_2, \ldots, q_N)$ with $q_{i,j} \in \mathbb{R}, j = 1, \ldots, d$, is called the configuration space and the space $\Gamma = \mathbb{R}^{dN} \times \mathbb{R}^{dN} = \mathbb{R}^{2dN}$ of the variables $\{q, p\}$ is called the phase space which characterizes the state of the system and the observables, are given by functions defined on the phase space. The Hamiltonian $H: \Gamma \to \mathbb{R}$ of the above system is defined by the observable

$$H(p,q) = \sum_{i=1}^{N} \frac{p_i^2}{2m_i} + V(q_1, q_2, \ldots, q_N)$$

which coincides with the sum of the kinetic and potential energies, traditionally denoted $T$ and $V$, respectively. The equations of motion read for all $i = 1, \ldots, N$ as

$$\dot{q}_i = \frac{\partial}{\partial p_i} H(q,p)$$

$$\dot{p}_i = -\frac{\partial}{\partial q_i} H(q,p)$$

$$(q(0), p(0)) = (q^0, p^0)$$

Where dot denote the derivative with respect to time and $\{q(t), p(t)\}$ are solutions with initial conditions $\{q(0), p(0)\}$. Assuming the Hamiltonian is time independent, the time evolution of any observable $B: \Gamma \to \mathbb{R}$ defined on phase space is governed by the Liouville equation

$$\frac{d}{dt} B_t(q,p) = L_H B_t(q,p)$$
$$B_0(q,p) = B(q,p)$$

where the linear operator $L_H$ is given by

$$L_H = \nabla_q \cdot \nabla_p H(q,p) - \nabla_p \cdot \nabla_q H(q,p)$$

Therefore, the formal solution given by

$$B_t(q,p) = e^{tL_H} B_0(q,p)$$

Consider a random variable $\theta \in \mathbb{R}^D$ with target density $\wp(\theta)$, in HMC scheme the auxiliary variable $p \in \mathbb{R}^D$ with factorized joint density

$$\wp(\theta, p) = \wp(\theta)\wp(p) = \wp(\theta)\mathcal{N}(p|0, M)$$

is introduced. In order to simplify the HMC methodology we denote the logarithm of target density by $\mathcal{L}(\theta) = \log\{\wp(\theta)\}$, furthermore to allow the use of deterministic dynamical transition and stochastic sampling by simulating the Hamiltonian dynamics the negative joint log density defined as

$$H(\theta, p) = -\mathcal{L}(\theta) + \frac{1}{2}\log\{(2\pi)^D |M|\} + \frac{1}{2} p^T M^{-1} p$$

Where $H$ is Hamiltonian and describe the sum of potential energy function $-\mathcal{L}(\theta)$ and the kinetic energy term $p^T M^{-1} p$ with momentum variable $p$ and covariance matrix $M$ as a mass matrix. Therefore, the dynamic system as given by Hamilton's equations

$$\frac{d\theta}{dt} = \frac{\partial H}{\partial p} = M^{-1} p$$

$$\frac{dp}{dt} = -\frac{\partial H}{\partial \theta} = \nabla_\theta \mathcal{L}(\theta)$$

The overall HMC sampling scheme from the invariant density $\wp(\theta)$ can be considered as

$$p^{n+1}|\theta^n \sim \wp(p^{n+1}|\theta^n) = \wp(p^{n+1}) = \mathcal{N}(p^{n+1}|0, M)$$

$$\theta^{n+1}|p^{n+1} \sim \wp(\theta^{n+1}|p^{n+1})$$

where samples of $\theta^{n+1}$ from $\wp(\theta^{n+1}|p^{n+1})$ are obtained by running the numerical integrator from initial values of $p^{n+1}$ and $\theta^n$ for a certain number of steps ($\ell$) to give a proposed moves to a new $\theta^\star$ and $p^\star$ with acceptance probability of

$$min[1, \exp(H(\theta^n, p^{n+1}) - H(\theta^*, p^*))]$$

An efficient probability space exploration can achieve by large $\ell$ and proper choice of $M$. As already mentioned, it is unclear how to select/ tune the entries of $M$ in automated manner to obtain the acceptable performance of HMC methods. Major step forward is RHMC scheme where the position dependent Fisher information matrix $G(x) = -\nabla_q^2 \mathcal{L}$ was introduced to play the role of $M$. However, it can be argued that the $G(x)$ is problem dependent and the calculation of metric tensor derivatives to be computationally too costly to implement. Under the Hamiltonian dynamics in HMC scheme, the proposal distribution $q(\theta^*|\theta)$ which drives the Markov chain takes the form of a classic random walk spread ballistically in probability space hence leads to low acceptance rate and highly correlated samples. It then follows that the automatic adaptation of pre-conditioned mass matrix can also cause localization where the posterior distribution is highly correlated. The localization due to the intrinsic linearity of unitary evolution of Hamiltonian system confirms this intuition.

The Quantum description of a above classical system is given by a set of postulates as

- The phase space $\Gamma$ is replaced by a Hilbert space $\mathcal{H} = L^2(\mathbb{R}^{dN}), \mathbb{R}^{dN}$ whose scalar product given by $\langle *|*\rangle$. The state of the system is characterized by a complex valued wave function $\psi(q) \text{in} L^2(\mathbb{R}^{dN})$.
- The observables are given by self-adjoint linear operators on $\mathcal{H}$.
- The result of the measure of an observable $B$ on the quantum system given by $\psi \in H$, is an element $b \in R$ of the spectrum $\sigma(B)$ of the self-adjoint operator $B$. Moreover, the probability to obtain an element in $(b1, b2]$ as the result of this measure on the state $\psi$ is given by $\mathcal{P}_\psi(B \in (b_1, b_2]) = ||P_B((b_1, b_2])\psi||^2$. Where $P_B(I)$ denotes the spectral projector of the operator $B$ on the set $I \subset \mathbb{R}$. Hence the expectation value of $B$ in $\psi$ can be written as $\mathbb{E}_{\psi(t)}(B) = \int_{\sigma(B)} b||P(db)\psi||^2 = \int_{\sigma(B)} b\langle\psi/P(db)\psi\rangle = \langle\psi/B\psi\rangle$
- The time evolution of the system is determined by its Hamiltonian $H$ according to the Heisenberg equation in the space of self-adjoint operators on $\mathcal{H}$ given as $i\hbar \frac{d}{dt} B(t) = -[H, B(t)], B(0) = B$

For example, a celebrated Hamiltonian which play a prominent role in quantum mechanics is quantum harmonic oscillator given as

$$H = \frac{\hat{p}^2}{2m} + \frac{1}{2}m\omega^2 \hat{q}^2$$

Where $\hat{p}$ is momentum operator given by $\hat{p} = -i\hbar \frac{\partial}{\partial q}$, $m$ is the particle's mass, $\omega = \sqrt{k/m}$ is angular frequency of the oscillator, $k$ is the force constant and $\hat{q}$ is position operator. One may write the time-independent Schrödinger equation as $\hat{H}|\psi\rangle = E|\psi\rangle$, where $E$ is time-independent eigenvalue and solution $|\psi\rangle$ denotes the eigenstate. We give here, some heuristics behind the formal definition of state (or mixed state) in quantum mechanics supported by given postulates which will be used later on. Before sketch the problem in the quantum mechanical framework it might be therefore useful to introduce some important topological and geometric properties of the space of density operators on a finite dimensional Hilbert space, that will be useful in order to better appreciate the challenges associated with probability measures on the set of density matrices. [10-11] In general, a complex $n \times n$ matrix $\rho$ is a density matrix if it has the following properties

- Hermitian: $\rho = \rho^\dagger$
- Positive: $\rho \geq 0$
- Normalized: $\text{Tr}\rho = 1$
- expectation value of operator $\hat{s}$ is given by $\langle \hat{s} \rangle = \text{tr}(\hat{\rho}(\tau)\hat{s})$

From now we consider $\mathcal{H}$ to be a complex $d$-dimensional Hilbert space, $\mathcal{M}$ be the space of density operators and $\hbar(\mathcal{H})$ be the space of Hermitian operators on $\mathcal{H}$. The set of density matrices is a convex set sitting in the vector space of Hermitian matrices and will be denoted $\mathcal{M}^{(N)}$ and its pure states form a complex projective space which obeying $\rho^2 = \rho$. Any density matrix which is diagonal in a given basis $\{|\psi_i\rangle\}$ can be expressed as eigen ensemble form as

$$\hat{\rho}(\tau) = \sum_{i=1}^{N} \lambda_i |\psi_i(\tau)\rangle\langle\psi_i(t)|$$
$$\sum_{i=1}^{N} \lambda_i = 1$$

Where $\lambda_i$ represent the probabilities(eigenvalue) for each quantum states $|\psi_i(\tau)\rangle$ (eigenstate). We note that the off-diagonal entries of $\rho(\tau)$ represent the individual probability flows while diagonal elements account for the total probability density outflow. A general representation of $\hat{\rho}(\tau)$ is $n \times n$ matrix in the basis of states $\{|1\rangle, ..., |N\rangle\}$ with elements $\rho_{ij}(\tau) = \langle i|\hat{\rho}(\tau)|j\rangle$ and the corresponding master equation under Lindblad form can be written as

$$\frac{d\rho_S(t)}{dt} = \mathcal{L}[\rho_S] = \sum_k -i[H, \rho_S] - \frac{1}{2}L_k^\dagger L_k \rho_S - \frac{1}{2}\rho_S L_k^\dagger L_k + L_k L_k^\dagger$$

Where $H$ is the Hamiltonian operator, $L_k$ is an arbitrary orthonormal basis of the operators on Hilbert space $\mathcal{H}_S$. From elementary quantum mechanics we all know that there are natural notions of probability amplitude and transition probability in the Hilbert space context. Suppose with a given two pure states $|\psi_1\rangle$ and $|\psi_2\rangle$ as a unit vector in a complex Hilbert space $\mathcal{H}$, then the probability amplitude for the system in state $|\psi_1\rangle$ to be found in state $|\psi_2\rangle$ is equal to $\langle\psi_2|\psi_1\rangle$ with the corresponding probability $|\langle\psi_2|\psi_1\rangle|^2$. Therefore, a more precise statements is that the geometry of projective Hilbert space which carries a natural metric structure the Fubini-Study metric is encoded in space of unit rank projector [8-9]. Hence a unit rank projector, representing a pure state is a special case of a density operator $\hat{\rho}(\tau)$ or mixed quantum states. Furthermore, considering the operator characterization of $\hat{\rho}(\tau)$ give a general expression of quantum Fisher information matrix as

$$\mathcal{J}_{\text{QFI}} = \sum_{i=1}^{N} 4\rho_i \langle\Delta^2 H\rangle_i - \sum_{i \neq j} \frac{8\rho_i\rho_j}{\rho_i + \rho_j}|\langle\psi_i|H|\psi_j\rangle|^2$$

Where $\rho_i$ and $|\psi_i\rangle$ are $i^{\text{th}}$ eigen value and eigen state of $\hat{\rho}(\tau)$ and

$$H := \iota\,(\partial_\theta U^\dagger)U$$

$$U = \exp(-\iota\tau\mathcal{L})$$

In order to identify the exponential representation of tangent vectors $\dot{\rho}$ at $\rho$ with traceless Hermitian operators we can use a real-valued inner products field $\kappa$ on $\hbar(\mathcal{H})$ which is smoothly parameterized by the density operators. Example of such filed include the symmetric generalized covariance defined as

$$\kappa_\rho(A,B) = \frac{1}{2}\mathrm{tr}(\rho A, B)$$

Where the curly bracket is the skew-commutator $\{A,B\} = AB + BA$ and the metric associated with $\kappa$ is quantum Fisher information metric $g_F$ defined as

$$g_F(\dot{\rho}_1, \dot{\rho}_2) = \frac{1}{2}\mathrm{tr}\,(\rho\{L^{\mathrm{BKM}}_{\dot{\rho}_1}, L^{\mathrm{BKM}}_{\dot{\rho}_2}\})$$

Where $L^{\mathrm{BKM}}_{\dot{\rho}_2}$ is Bogolubov's logarithmic derivative of $\dot{\rho}$ at $\rho$ defined as

$$L^{\mathrm{BKM}}_{\dot{\rho}} = \frac{d}{dt}\log\rho_t|_{t=0}$$

Where $\rho_t$ is a curve extending from $\rho$ with initial velocity $\dot{\rho}$. Shortly speaking, the Bogolubov-Kubo-Mori metric (BKM) can be derived from quantum version of Kullback-Leibler divergence (quantum relative entropy) as

$$g^{\mathrm{KMB}}_{jk}(\theta_0) = \frac{\partial^2}{\partial\theta_j\partial\theta_k}|_{\theta\to\theta_0}D_{\mathrm{KL}}(\rho_{\theta_0}||\rho_\theta)$$

In this context the quantum relative entropy could be defined as follows

$$D_{\mathrm{KL}}(\rho||\sigma) = \mathrm{tr}\,(\rho\,(\log\rho - \log\sigma)), \rho,\sigma \in \mathcal{S}(\mathcal{H})$$

Where $\mathcal{S}(\mathcal{H})$ is the set of all density operators on $\mathcal{H}$ such that $\mathcal{S}(\mathcal{H}) = \{\rho: \mathrm{tr}(\rho) = 1, \rho \geq 0\}$. In view of such considerable interest and challenges associated with probability measures on the space of density matrices, now we turn to the main subject of the paper which is the parametric formulation of statistical inference theory in quantum mechanical terms. Apart from the new insights it involves restatement of Markov jump proposals in probability space exploration in Bayesian perspective with quantum mechanical density operators. Although the main context will incorporate the quantum jump proposals for Hamiltonian Monte Carlo scheme, the application of these concept is not limited there to. So far, we introduce some useful terminology and we also described the structure of quantum mechanical ensembles, now in order to tackle the problem of posterior sampling in statistical inference theory we define the probability measures on the set of density matrices through a metric. This corresponds to stochastic quantum trace preserving maps, which provide a kind of stroboscopic time evolution in a given space. Indeed, the simplest example is a unitary transformation. Consider a family of quantum states $\rho_\theta$ which are defined on a given $\mathcal{H}$ parametrized by $\theta$ on a d-dimensional manifold $\mathcal{M}$ where the states are obtained from a given initial state $\rho_0$ by the action of unitary operation as

$$\rho_\theta = U_\theta(\rho_0) = U_\theta \rho_0 U_\theta^\dagger$$

$$U_\theta = \exp\{-i\theta H\}$$

By expanding the initial state in its eigen-basis

$$\rho_0 = \sum \lambda_i|\phi_i\rangle\langle\phi_i|$$

$$\rho_\theta = \sum_n \lambda_n|\psi_n\rangle\langle\psi_n|$$

with $|\psi_n\rangle = U_\lambda|\phi_n\rangle$.
Consequently, we have

$$\partial_\theta \rho_\theta = iU_\theta[H,\rho_0]U_\theta^\dagger$$

To gain insight into dissipative processes and incoherent excitations, we now consider the time evolution of an arbitrary density operator using the structural theorem of Lindblad formalism. Let's recall the dynamics of open driven and dissipative quantum systems in Lindblad formalism where non-unitary energy fluctuations put a lot of thermo-dynamical information into the matrix elements of $\rho$ operator. Therefore, the resulting time evolution can be divided into a coherent time evolution governed by a non-Hermitian Hamiltonian operator interrupted by instantaneous jump operators and the consequent gain (localization in position space) in knowledge about the system. Let $\mathcal{H}$ be a complex, separable Hilbert space, $(\Omega, \mathcal{F}, (\mathcal{F}_t), \mathbb{Q})$ be a stochastic basis where d-dimensional continuous Wiener process is defined. we denote $\mathcal{T}(\mathcal{H})$ be the trace class on $\mathcal{H}$, $\mathcal{S}(\mathcal{H})$ the subset of the statistical operators and $\mathcal{L}(\mathcal{H})$ be the space of the linear bounded operators on $\mathcal{H}$. The linear stochastic Schrodinger equation (LSSE) define as

$$d\psi(t) = K(t)\psi(t)\,dt + \sum_{j=1}^{d} R_j(t)\psi(t)\,dW_j(t),$$

$$\psi(0) = \psi_0 \in L^2(\Omega, \mathcal{F}_0, \mathbb{Q}; \mathcal{H})$$

where the drift term is

$$K(t) = -iH(t) - \frac{1}{2}\sum_{j=1}^{d} R_j(t)^* R_j(t)$$

The coefficient $H(t), R_j(t)$ are stochastic bounded operators as $H(t) = H(t)^*$ on $(\Omega, \mathcal{F}, (\mathcal{F}_t), \mathbb{Q})$ where $\mathbb{Q}$ is reference probability measure. Moreover, $\forall T > 0$

$$\int_0^T \mathbb{E}_\mathbb{Q}[||H(t)||]\,dt \langle +\infty,$$

$$\mathbb{E}_\mathbb{Q}[\exp\{2\sum_{j=1}^{d}\int_0^T ||R_j(t)^2||\,dt\}]\langle +\infty$$

which implies

$$\|\psi(t)\|^2 = \|\psi_0\|^2 \exp\left\{\sum_j \left[\int_0^t m_j(s)\, dW_j(s) - \frac{1}{2}\int_0^t m_j(s)^2\, ds\right]\right\},$$

$$m_j(t) := 2\text{Re}\langle \hat{\psi}(t) | R_j(t) \hat{\psi}(t) \rangle,$$

$$\hat{\psi}(t) := \begin{cases} \psi(t)/\|\psi(t)\| & \text{if } \|\psi(t)\| \neq 0, \\ v & \text{if } \|\psi\| = 0 \end{cases}$$

Let us define $\alpha \in \mathcal{L}(\mathcal{H})$ and the $\mathcal{T}(\mathcal{H})$ process as $\sigma(t) := |\psi(t)\rangle\langle\psi(t)|$, then by applying Ito formula to $\langle \psi(t) | \alpha \psi(t)\rangle$ we can get

$$d\sigma(t) = \mathcal{L}(t)[\sigma(t)]\, dt + \sum_{j=1}^{d} \mathcal{R}_j(t)[\sigma(t)]\, dW_j(t)$$

$$\mathcal{R}_j(t)[\rho] := R_j(t)\rho + \rho R_j(t)^*$$

$$\mathcal{L}(t)[\rho] = i[H(t), \rho]$$
$$+ \sum_{j=1}^{d} \left(R_j(t)\rho R_j(t)^* - \frac{1}{2}\{R_j(t)^* R_j(t), \rho\}\right)$$

Where $\mathcal{L}$ is stochastic Liouville operator. By assuming

$$\sup_{\omega \in \Omega} \int_0^t \left\|\sum_{j=1}^{d} R_j(s,\omega)^* R_j(s,\omega)\right\| ds < +\infty$$

$\hat{\psi}(t)$ satisfies the non-linear stochastic Schrödinger equation

$$d\hat{\psi}(t) = \sum_j [R_j(t) - \text{Re}\, \eta_j(t, \hat{\psi}(t))]\hat{\psi}(t)\, d\widehat{W}_j(t)$$
$$+ K(t)\hat{\psi}(t)\, dt$$
$$+ \sum_j \left[(\text{Re}\, \eta_j(t, \hat{\psi}(t))) R_j(t)\right.$$
$$\left. - \frac{1}{2}(\text{Re}\, \eta_j(t, \hat{\psi}(t)))^2\right]\hat{\psi}(t)\, dt$$

Where $\eta_j(t, x) := \langle x | R_j(t) x \rangle, \forall t \in [0, +\infty], j = 1,\ldots,d, x \in \mathcal{H}, \widehat{W}_j(t) := W_j(t) - \int_0^t m_j(s)\, ds, j = 1,\ldots,d, t \in [0,T]$ is standard Wiener process. Note that $\text{tr}\{\alpha\eta(t)\} = \mathbb{E}_{\mathbb{Q}}[\langle \psi(t) | \alpha \psi(t)\rangle] \forall \alpha \in \mathcal{L}(\mathcal{H})$ satisfies

$$\eta(t) = \varrho_0 + \int_0^t \mathbb{E}_{\mathbb{Q}}[\mathcal{L}(s)[\sigma(s)]]\, ds$$

then we have the following non-linear equation

$$\begin{cases} d\varrho(t) = \mathcal{L}(t)[\varrho(t)]\, dt + \sum_{j=1}^{d} \mathcal{R}_j(t)[\varrho(t)] - v_j(t)\varrho(t)\, d\widehat{W}_j(t) \\ \varrho(0) = \varrho_0 \end{cases}$$

Where

$$\varrho(t) = \frac{\sigma(t)}{\text{tr}\,\sigma(t)}$$

$$v_j(t) := \text{tr}(R_j(t) + R_j(t)^*)\varrho(t)$$

$$\widehat{W}_j(t) := W_j(t) - \int_0^t v_j(s)\, ds, \forall j = 1,\ldots,d$$

In order to adopt the memory strategy into LSSE by driven noise, let us consider bounded drift and diffusion operators $A, B$ on $\mathcal{H}$ as

$$d\psi(t) = A\psi(t)\, dt + B\psi(t)\, dX(t)$$

Where $X(t)$ is stationary Ornstein-Uhlenbeck process, as

$$tX(t) = e^{-\gamma\tau}Z + \int_0^t e^{-\gamma(\tau-s)}\, dW(s), \gamma > 0$$

where $Z$ is $\mathcal{F}_0$-measurable gaussian random variable with mean zero and variance $1/(2\gamma)$. Therefore, its straightforward to write

$$d\psi(t) = (A - \gamma X(t)B)\psi(t)\, dt + B\psi(t)\, dW(t)$$

now with

$$H_0 = H_0^* \in \mathcal{L}(\mathcal{H})$$
$$H(t) = H_0 - \gamma X(t) L$$
$$R(t) = -iL$$
$$K(t) = -i(H_0 - \gamma X(t)L) - \frac{1}{2}L^2,$$

$$d\psi(t) = \left[-i(H_0 - \gamma X(t) L) - \frac{1}{2}L^2\right]\psi(t)\, dt$$
$$- iL\psi(t)\, dW(t)$$

$$\psi(t) = \overleftarrow{T} \exp - i \int_0^t (H_0 - \gamma X(t) L)\, ds - i \int_0^t L\, dW(s)\, \psi_0$$

Furthermore, the evolution of the corresponding density matrices $\sigma(t) = |\psi(t)\rangle\langle\psi(t)|$ satisfies

$$d\sigma(t) = -i[H_0 - \gamma X(t)L, \sigma(t)]\, dt - i[L, \sigma(t)]\, dW(t)$$
$$- \frac{1}{2}[L, [L, \sigma(t)]]\, dt$$

hence the evolution of the mean given as

$$\frac{d}{dt}\eta(t) = -i[H_0, \eta(t)] - \frac{1}{2}[L, [L, \eta(t)]]$$
$$+ i\gamma[L, \mathbb{E}_{\mathbb{Q}}[X(t)\sigma(t)]]$$

In order to generalize the positivity of master equation using the Nakajima-Zwanzig projection technique let us consider $\mathcal{P}[\ldots] := \mathbb{E}_{\mathbb{Q}}[\ldots], \mathcal{Q} = \mathbb{I} - \mathcal{P}$, then we have

$$\eta(t) = \mathcal{P}[\sigma(t)]$$
$$\sigma_\perp(t) := \mathcal{Q}[\sigma(t)] = \sigma(t) - \eta(t),$$
$$\mathcal{L}_M(t) := \mathbb{E}_{\mathbb{Q}}[\mathcal{L}(t)],$$
$$\Delta\mathcal{L}(t) := \mathcal{L}(t) - \mathcal{L}_M(t)$$

By using the projection operators and Ito formula we get

$$\dot{\eta}(t) = \mathcal{L}_M(t)[\eta(t)] + \mathcal{P}\circ\Delta\mathcal{L}(t)[\sigma_\perp(t)],$$

$$d\sigma_\perp(t) = \mathcal{Q}\circ\mathcal{L}(t)[\sigma_\perp(t)]\, dt + \sum_{j=1}^{d} \mathcal{R}_j(t)[\sigma_\perp(t)]\, dW_j(t)$$
$$+ \mathcal{Q}\circ\mathcal{L}(t)[\eta(t)]\, dt$$
$$+ \sum_{j=1}^{d} \mathcal{R}_j(t)[\eta(t)]\, dW_j(t)$$

$$\sigma_\perp(t) = Q o \mathcal{V}(t,0)[\sigma_\perp(0)]$$
$$+ \int_0^t Q o \mathcal{V}(t,s) o (\mathcal{L}(s) - \sum_j \mathcal{R}_j(s)^2)[\eta(s)]\, ds$$
$$+ Q o \mathcal{V}(t,0)\left[\sum_{j=1}^d \int_0^t \mathcal{V}(s,0)^{-1} o \mathcal{R}_j(s)[\eta(s)]\, dW_j(s)\right]$$

which propagator $\mathcal{V}(t,r)$ satisfies
$$\mathcal{V}(t,r) = \mathbb{I} + \int_r^t ds\, \mathcal{L}(s) o \mathcal{V}(s,r)$$
$$+ \sum_{j=1}^d \int_r^t dW_j(s)\, \mathcal{R}_j(s) o \mathcal{V}(s,r)$$

Then by introducing the projectors we get the master equation
$$\dot\eta(t) = J(t) + \mathcal{L}_M(t)[\eta(t)] + \int_0^t \mathcal{K}(t,s)[\eta(s)]\, ds$$
$$+ \mathbb{E}_\mathbb{Q}\left[\Delta \mathcal{L}(t) o Q o \mathcal{V}(t,0)\left[\sum_{j=1}^d \int_0^t \mathcal{V}(s,0)^{-1} o \mathcal{R}_j(s)[\eta(s)]\, dW_j(s)\right]\right]$$

where inhomogeneous term and integral memory kernel can be written respectively as
$$J(t) := \mathbb{E}_\mathbb{Q}[\Delta \mathcal{L}(t) o Q o \mathcal{V}(t,0)[\sigma_\perp(0)]]$$
$$\mathcal{K}(t,s) := \mathbb{E}_\mathbb{Q}[\Delta \mathcal{L}(t) o Q o \mathcal{V}(t,s) o (\mathcal{L}(s) - \sum_j \mathcal{R}_j(s)^2)]$$

In order to numerically simulate the stochastic Schrodinger equation, we can approximate $\mathcal{V}(t,r)$ by
$$\sigma_\perp(0) = 0$$
$$\mathcal{V}_M(t,r) = \mathbb{I} + \int_r^t ds\, \mathcal{L}_M(s) o \mathcal{V}_M(s,r)$$
$$\dot\eta(t) \simeq \mathcal{L}_M(t)[\eta(t)] + \int_0^t \mathcal{K}_1(t,s)[\eta(s)]\, ds$$
$$+ \mathbb{E}_\mathbb{Q}\left[\Delta \mathcal{L}(t)\left[\sum_{j=1}^d \int_0^t \mathcal{V}_M(t,s) o \mathcal{R}_j(s)[\eta(s)]\, dW(s)\right]\right],$$
$$\mathcal{K}_1(t,s) := \mathbb{E}_\mathbb{Q}\left[\Delta \mathcal{L}(t) o \mathcal{V}_M(t,s) o \left(\Delta \mathcal{L}(t) - \sum_j \Delta \mathcal{R}_j^2(s)\right)\right]$$
$$\Delta \mathcal{R}_j^2(s) = \mathcal{R}_j(s)^2 - \mathbb{E}_\mathbb{Q}[\mathcal{R}_j(s)^2]$$

and the approximation of the non-Markovian master equation become
$$\frac{d\eta}{dt} \simeq -i[H_0, \eta(t)] - \frac{1}{2}[L,[L,\eta(t)]]$$
$$+ \frac{\gamma}{2}\int_0^t ds[L, e^{(\mathcal{L}_M - \gamma)(t-s)}[[L,\eta(s)]]]$$

In the virtue of above perspective, the probability measure on space of density matrix is Circular $\beta$-Ensemble(C$\beta$E) as
$$U(\tau + \delta\tau) = U(\tau)\exp(i\sqrt{\delta\tau}M(\tau))$$

Where $\delta\tau$ is infinitesimal and $M(\tau)$ is defined as real and imaginary parts of the density matrix elements. Then the joint-probability density for transitions to circular ensemble for arbitrary $\tau$ is constructed as follows: Let $M$ be a Hermitian matrix with complex elements and let $H$ be decomposed in terms of its eigenvalues and eigenvectors via $M = ULU^\dagger$, where $L$ is a diagonal matrix consisting of the eigenvalues of $M$, and U is a unitary matrix with complex elements consisting of the corresponding eigenvectors, therefore we have

$$(d\mathrm{M}) = \prod_{1\le j<k\le N}|\lambda_k - \lambda_j|^2 \bigwedge_{j=1}^N d\lambda_j(U^\dagger dU)$$

Where $d\mathrm{M}$ is volume form and from diagonalization formula $U = U_2\Theta U_2^\dagger$ we can write
$$U_2^\dagger dUU_2 = \delta U_2 \Theta - \Theta\delta U_2 + i\Theta d\theta$$
$$\delta U_2 = U_2^\dagger dU_2$$
$$\Theta =, \mathrm{diag}[e^{i\theta_j}]_{j=1,\ldots,N}$$

And $\theta$ is diagonal matrix with entries $\theta_{j=1,\ldots,N}$, therefore gives

$$(U^\dagger dU) = \prod_{1\le j<k\le N}|e^{i\theta_j} - e^{i\theta_k}|^2 \bigwedge_{j=1}^N d\theta_j(U_2^\dagger dU_2)$$

Where $(U_2^\dagger dU_2) = \bigwedge_{j<k}\delta u_{2jk}^r \delta u_{2jk}^i$ and each eigenvalue written as $\lambda_j = e^{i\theta_j}$. Therefore, the diagonalization of $\rho$ leads to the construction of a member of the circular unitary ensemble (CUE). So far as investigation goes the interpretation of stochastic evolution process based on quantum master equation arrived at the desired generalization of momentum proposal scheme for efficient Bayesian hierarchical modelling in the following form

$$p^{n+1}|\theta^n \sim \wp(p^{n+1}|\theta^n) = \wp(p^{n+1}) = \mathcal{N}(p^{n+1}|0, M|\rho)$$
$$\theta^{n+1}|p^{n+1} \sim \wp(\theta^{n+1}|p^{n+1})$$

A remarkable feature of proposed scheme which framed in terms of the density matrix $\hat\rho(\tau)$ is a transition from initially ballistic quantum walk Hamiltonian to incoherent scattering described by Lindblad operators from open quantum system theory where the coherent evolution is stochastically perturbed by the action of jump operator. Having found the proposal mechanism one can numerically integrate the Hamiltonian system by solving the continuous time derivatives in order to get the new state in the probability space. We have therefore shown the transition from the dressed state basis to intrinsic decoherence characteristics can lead to gain in information about the system dynamics. However, we do not intend to elaborate further on the mathematical part of the theory. Instead, we would like to show from specific examples this observation will resolve the issues associated with HMC schemes.

### III. NUMERICAL EXPERIMENTS

In this section we report the experimental results of performance sketch divided into two benchmarks: (1) The first part reports a simulation study based on ill-conditioned $\mathcal{D}$-dimensional Gaussian distribution, with $\mathcal{D}$ =10, 50, 100, 150, 200, 500. In particular we consider, centered $\mathcal{D}$-dimensional Gaussian with covariance eigen values of $[\frac{1}{10}, 1, \ldots, 10^{64}]$ (eigen values ranges corresponding to each $\mathcal{D}$-dimensional covariance matrix). In order to explore the convergence of proposed method for the first benchmark we use Kullback–Leibler divergence which is a measure used to calculate the difference between two probability distribution. Suppose we have two multivariate normal distribution with mean $\mu$, $\hat{\mu}$ and positive definite covariance matrix $\Sigma$, $\hat{\Sigma}$ (where the over bar indicates the estimated distribution) in k dimension. In our first experiment KL divergence between target $\mathcal{N}(\mu, \Sigma)$ and estimated density $\widehat{\mathcal{N}}(\hat{\mu}, \hat{\Sigma})$ denoted by $D_{KL}(\mathcal{N}(\mu, \Sigma) \parallel \widehat{\mathcal{N}}(\hat{\mu}, \hat{\Sigma}))$ can be obtained with appropriate substitution parameters in the equation below

$$D_{KL}(\mathcal{N}(\mu, \Sigma) \parallel \widehat{\mathcal{N}}(\hat{\mu}, \hat{\Sigma})) = \frac{1}{2}(\text{tr}(\Sigma^{-1} \cdot \hat{\Sigma}) + (\mu - \hat{\mu})^T \cdot \Sigma^{-1} \cdot (\mu - \hat{\mu}) - k + \ln(\frac{\det \Sigma}{\det \hat{\Sigma}}))$$

Visualization of KL divergence and computational cost for first benchmark in given in Fig. (1) and table (1). In this example we consider 100 chains of parallel replicas (HMC chains) with $10^7$ time steps, hence the given results averaged over this configuration properties.

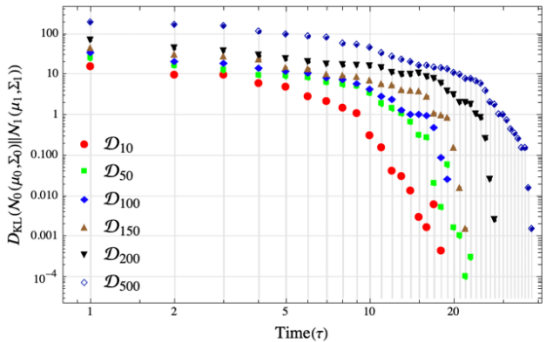

**Fig 1.** Visualization of convergence diagnostic tool ($D_{KL}$) for kernel density estimation of multi normal distribution. $\mathcal{D}$ indicate the dimensionality of target density. Plotted values are divergence rate along the PDMC iteration step.

| Dimension | $D_{KL}(\mathcal{N}(\mu, \Sigma) \parallel \widehat{\mathcal{N}}(\hat{\mu}, \hat{\Sigma}))$ | $CPU_{\langle s \rangle}$ |
|---|---|---|
| $\mathcal{D}_{10}$ | 0.00004 | 270.0 |
| $\mathcal{D}_{50}$ | 0.00072 | 2300.0 |
| $\mathcal{D}_{150}$ | 0.0146 | 12680.0 |
| $\mathcal{D}_{200}$ | 0.0234 | 18090.0 |
| $\mathcal{D}_{500}$ | 0.04563 | 32940.0 |

**Table 1.** Density Estimation convergence rate. ($\mathcal{D}$-Dimensional Gaussian distribution). $CPU_{(s)}$ denotes the average CPU time in second for the whole run in the first experiment.

For the second test we consider the estimation problem of first 150 smallest magnitude eigenvalue problem for Airy operator given as $\mathcal{A} = -\partial_x^2 + x$, endowed with homogeneous Dirichlet boundary condition at x = 0, in an interval, where the eigenvalues are roots of a transcendental equation.

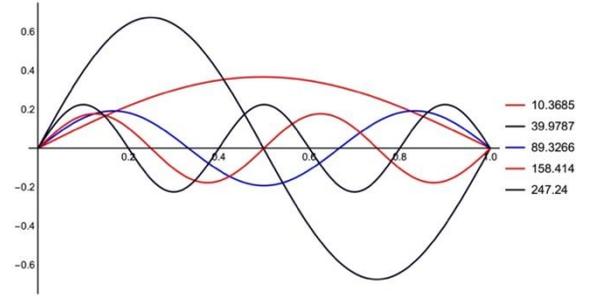

**Fig 2.** Numerically computed first five smallest magnitude eigenvalue problem associated with Airy operator.

Therefore, a sequence of process observations (t) is modeled as $y(t) = x(t) + \epsilon(t)$ where $\epsilon(t)$ defines an appropriate multivariate gaussian process with zero mean and variance $\sigma^2_i$ for each state $x(t)$. The posterior density read as

$$\pi(\theta|Y, x_0, \sigma) \propto \pi(\theta) \prod_n \mathcal{N}(Yn|X(\theta, x_0)_n, \Sigma^{-1})$$

In this case the integration of HMC is carried out with a symplectic method with a relatively large number of steps, and the numerical approximations for the estimated eigenvalues in contrast to exact values yield an error around the order of unit roundoff in IEEE double-precision arithmetic that remains bounded. The following figure shows the absolute estimated error versus exact eigen values.

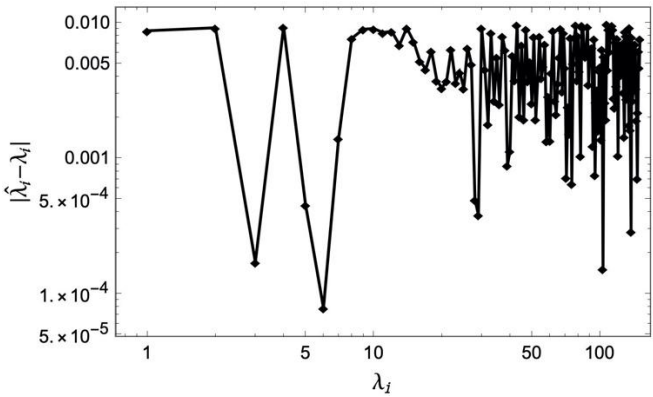

**Fig 3.** Error plot for estimated eigen values $\hat{\lambda}_i$ versus exact $\lambda_i$ for $i{\rm h}$ eigen problem associated with Airy operator in log scale.

For the purpose of second sampling benchmark, we collect 15000 posterior data points within 10 simulations. The performance summary current experiment is given as Table 2. The results of all numerical and symbolic demonstration (Matrix Exponential) were performed on Late 2018 MacBook pro with 16GB RAM running Mathematica programming language, same manner for sampling procedure. Together with an efficient class of symplectic partitioned Runge-Kutta method such as NDSolve, Mathematica greatly streamlines the development of our efficient code.

## IV. Conclusion

We have discussed the density matrix representation of Fisher information by considering the operator characterization of Lindblad formalism. Necessary condition for obtaining the relevant analytical expression of quantum Fisher information is determined. Furthermore, regarding the contribution made by this generalization in the language of inferential statistics, explicit algorithm for density estimation of arbitrary order is also provided. Apart from theoretical framework the experimental results confirm the usability, performance and accuracy of proposed method. Lastly some details of nonparametric density estimation and its computational aspects in the context of Bayesian statistic demonstrated.